\documentclass[a4paper]{article}

\usepackage{lmodern}
\usepackage[T1]{fontenc}
\usepackage[utf8]{inputenc}
\usepackage{amssymb,amsmath}
\usepackage{todonotes}
\usepackage{caption, subcaption}
\usepackage{url}

\usepackage{hyperref}
\hypersetup{unicode=true,
	pdfborder={0 0 0},
	breaklinks=true}
\usepackage{longtable,booktabs}
\IfFileExists{parskip.sty}{%
	\usepackage{parskip}
}{% else
\setlength{\parindent}{0pt}
\setlength{\parskip}{6pt plus 2pt minus 1pt}
}
\title{OpenPodcar: an Open Source Vehicle \\for Self-Driving Car Research}
\usepackage{authblk} 
\author[1,2]{Fanta Camara}
\author[2]{Chris Waltham}
\author[2]{\\Grey Churchill} %David has recently changed name (and gender, to they/them)
\author[1,2]{Charles Fox} 
\affil[1]{Institute for Transport Studies, University of Leeds, UK} 
\affil[2]{School of Computer Science, University of Lincoln, UK}

\begin{document}
	\maketitle

	\begin{abstract}
		OpenPodcar is a low-cost, open source hardware and software, autonomous vehicle research platform based on an off-the-shelf, hard-canopy, mobility scooter donor vehicle. Hardware and software build instructions are provided to convert the donor vehicle into a low-cost and fully autonomous platform. The open platform consists of (a) hardware components: CAD designs, bill of materials, and build instructions; (b) Arduino, ROS and Gazebo control and simulation software files which provide standard ROS interfaces and simulation of the vehicle; and (c) higher-level ROS software implementations and configurations of standard robot autonomous planning and control, including the move\_base interface with Timed-Elastic-Band planner which enacts commands to drive the vehicle from a current to a desired pose around obstacles. The vehicle is large enough to transport a human passenger or similar load at speeds up to 15km/h, for example for use as a last-mile autonomous taxi service or to transport delivery containers similarly around a city center. It is small and safe enough to be parked in a standard research lab and be used for realistic human-vehicle interaction studies. System build cost from new components is around USD7,000 in total in 2022. OpenPodcar thus provides a good balance between real world utility, safety, cost and research convenience.
	\end{abstract}
	
	\begin{longtable}[]{@{}l@{}}
		\begin{minipage}[t]{0.97\columnwidth}\raggedright\strut

			\subsection*{Metadata Overview}\label{h.akaipbqoqfs8}
			
			Main design files: \url{https://github.com/OpenPodcar/OpenPodcar}
			
			Target group: researchers and hobbyists interested in autonomous vehicle research and robotics. 
			
			Skills required: Mechanical assembly – intermediate (drilling steel); electrical assembly – intermediate (PCB soldering); Software – easy (Linux command line).
			
			Replication: The current OpenPodcar is being used by some of the authors for human-robot interaction experiments and a second copy will be built from the documentation to improve its accuracy. The design is currently being forked for a courier-type manually-driven platform by a commercial UK vehicle manufacturer.

			\subsection*{Keywords}\label{h.kdz351yp7g7c}
			
			{Autonomous vehicle, automation, self-driving car, mobility scooter, open source platform.}
			
			\strut\end{minipage}\tabularnewline
		\bottomrule
	\end{longtable}

	\section{(1) Overview}
	
	\subsection{Introduction}
	\label{h.pnj38xyr5dyy}
	
	Autonomous Vehicles (AVs, also known as `self-driving cars'), is a fast-moving research field in both academia and the industry. Open source software (OSS) for localisation, mapping and control of AVs is available \cite{kato2018autoware} but hardware vehicle platforms remain expensive and proprietary, making it difficult for researchers with low resources to develop algorithms or reproduce complete research systems. There is thus a need for a standard, low-cost, reproducible hardware platform, compatible with the standard open source software stack.
	
	Open source hardware (OSH) allows for more effective and accessible sharing and collaboration among researchers \cite{fisher2012open}. By combining OSH and OSS, a standard platform can be produced for use by all members of a research community, who may then reproduce each others work in full, and contribute their new research as functional system components rather than only as reports. Such platforms may evolve from research into development and real-world applications.
	
	To create an OSH platform for the autonomous vehicle research community, several requirements must be met: {\em low cost} and {\em easy to build} to enable the community to reproduce and use it; consumer levels of safety and reliability are not required, though {\em research standards of safety and reliability} are required; the system should be designed to enable {\em easy modification} so that it can be forked to operate with similar but different vehicles; the system should be physically {\em light-weight} to ease experimentation and reduce risks of damage, though large enough for {\em human transport} so that it can be used in real-world applications and in research requiring realistic interactions with other human road users \cite{camara2021evaluating, delucia2013effects}.
	
	\subsection{Related Systems}
	\label{related_systems}
	
	SMART \cite{pendleton2015autonomous} is a design to modify an existing donor golf cart vehicle for automation research, this is of a similar size and power to OpenPodcar. Similarly, iCab (Intelligent Campus Automobile) \cite{gomez2016ros} is research golf car with a ROS (Robot Operating System)-based architecture and that has been tested with Timed-Elastic Band planner \cite{marin-plaza2018global}. However, these vehicle designs are not open source hardware.
	
	Complete and built mechanical OSH designs for on-road, person-carrying cars exist, including PixBot \cite{pixmovingpixbot} and Tabby EVO \cite{openmotorstabby}. Building these full size cars is a large task for experts and may require dangerous processes such as welding, purchase of expensive components, and considerable storage space. OpenPodcar is based on a proprietary but commodity mobility scooter which is cheaper and easier to convert than performing these builds. 
	
	Several OSH RC-scale cars have been completed and built such as F1Tenth \cite{f1tenth}, AutoRally \cite{goldfain2019autorally}, BARC \cite{gonzales2018planning}, MIT Racecar \cite{mit}, MuSHR \cite{mushr}, \cite{nakamoto2019development}, and \cite{vincke2021open}. These platforms are not large enough to drive on public roads or to transport people or goods like OpenPodcar. Open Source Ecology (OSE) \cite{jakubowski2003open} is an ambitious programme of projects which ultimately aims to develop fully OSH vehicles including a car and tractor. OSE is optimised for reliability and for users in developing countries so it uses hydraulic power rather than electric as used in OpenPodcar. But its vehicle designs are not yet complete.
	
	Autoware \cite{kato2018autoware} is a heavyweight open source software project to construct a full ROS-based automation stack for on-road cars. Apollo \cite{apolloautoapollo} is an open source self-driving software stack and an open hardware interface which may be implemented on vehicles, as done in \cite{kessler2019bridging}. These systems could be software interfaced to run with OpenPodcar. 
	
	Some AV research can be done in simulation without the need for hardware, hence open source simulation platforms are widely available such as SUMMIT \cite{cai2020summit}, Gym-Duckietown \cite{chevalier-boisvert2018duckietown}, CARLA \cite{dosovitskiy2017carla}, DEEPDRIVE \cite{quiterdeepdrive}, LGSVL Simulator \cite{rong2020lgsvl}, AirSim\cite{shah2017airsim}, and FLOW \cite{wu2017flow}. The USA state of Georgia provides a level 3 open-source autonomous vehicle based on a Ford-Edge\cite{peachtreecuriosity}, which can be used {\em gratis} in their Peachtree Corners' Curiosity Lab smart city environment by researchers needing a vehicle but not wanting to build or buy one.
	
	\section{(2) Overall Implementation and Design}\label{h.1u7vph94gfbt}
	
	\subsection{Donor vehicle}
	
	A Pihsiang TE-889XLSN hard-canopy scooter (branded in UK as Shoprider Traverso, \cite{shoprider2016shoprider}) is used as a donor vehicle. It is an Ackermann-steered \cite{milliken1995race}, hard-canopy, electric mobility scooter. It is powered by two 12V batteries connected in series to provide 24V operating voltage and containing 75Ah. In its standard configuration, its steering is controlled by a human-operated loop handle bar. The speed and braking systems are both powered by an electric motor and an electric brake via the trans-axle assembly, controlled by an AC2 digital controller receiving different voltage signals to drive forward or brake. The manual speeding and braking systems are controlled by three buttons connected in series on the handle bar. A toggle switch in parallel with a resistor (10k$\Omega$) selects  speed mode from high (max 8mph) or low (max 4mph); a speed dial knob via a variable resistor (20k$\Omega$) sets a maximum limit speed within the mode. A throttle lever connected with a 5k$\Omega$ potentiometer is used to select the speed within the mode and limit.
	
	\subsection{Mechanical Modification for Steering}
	
\begin{figure}
		\centering
		\begin{subfigure}{0.45\textwidth}
			\centering
			\includegraphics[scale=0.29]{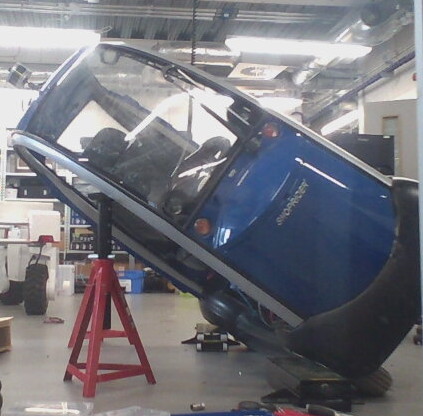}
			\caption{Tilting the vehicle using two axle stands, to enable access to the underside. (Note also lidar mounted to roof.)}
			\label{fig:axelStands}
		\end{subfigure}	
		\quad
		\begin{subfigure}{0.45\textwidth}
			\centering
			\includegraphics[scale=0.23]{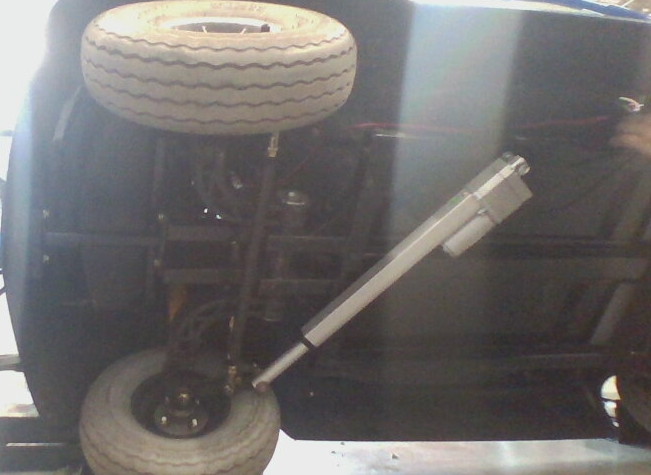}
			\caption{Underside with linear actuator added for steering.}
			\label{fig:actuatorMounted}
		\end{subfigure}
		\caption{Vehicle mechanical modification}
	\end{figure}

\begin{figure}[h]
		\includegraphics[width=\columnwidth]{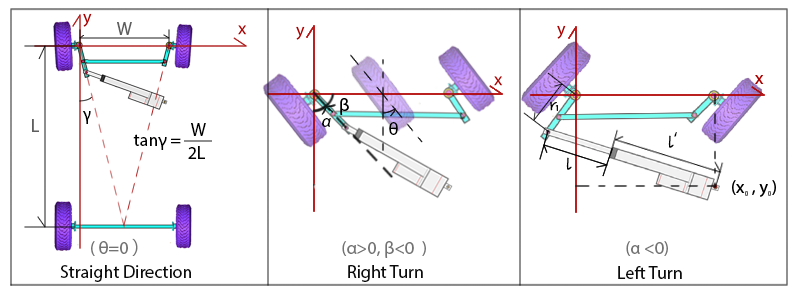}
		\caption{Underside view of front wheels' steering relationship including geometric coefficients}
		\label{fig:steering}
	\end{figure}

	To automate steering, a linear actuator (Gimson GLA750-P 12V DC) with position feedback is mounted between an anchor on the underside of the chassis and the car's front axle via bearings.  This actuator has a 8mm/s full load (750N) speed and 250mm stroke length (installation length is 390mm). To access the underside of the vehicle, two axle stands are used as shown in Fig. \ref{fig:axelStands}. There is an existing hole in the right front wheel axle. The linear actuator is mounted via a rear hole to the left side of the front chassis and connected through the front hole of the actuator with the hole in the car's right front wheel axle via bearings as shown in Figs. \ref{fig:actuatorMounted} and \ref{fig:steering}.

	\subsection{Electronics}
	
	\begin{figure*}[h]
		\includegraphics[width=\linewidth]{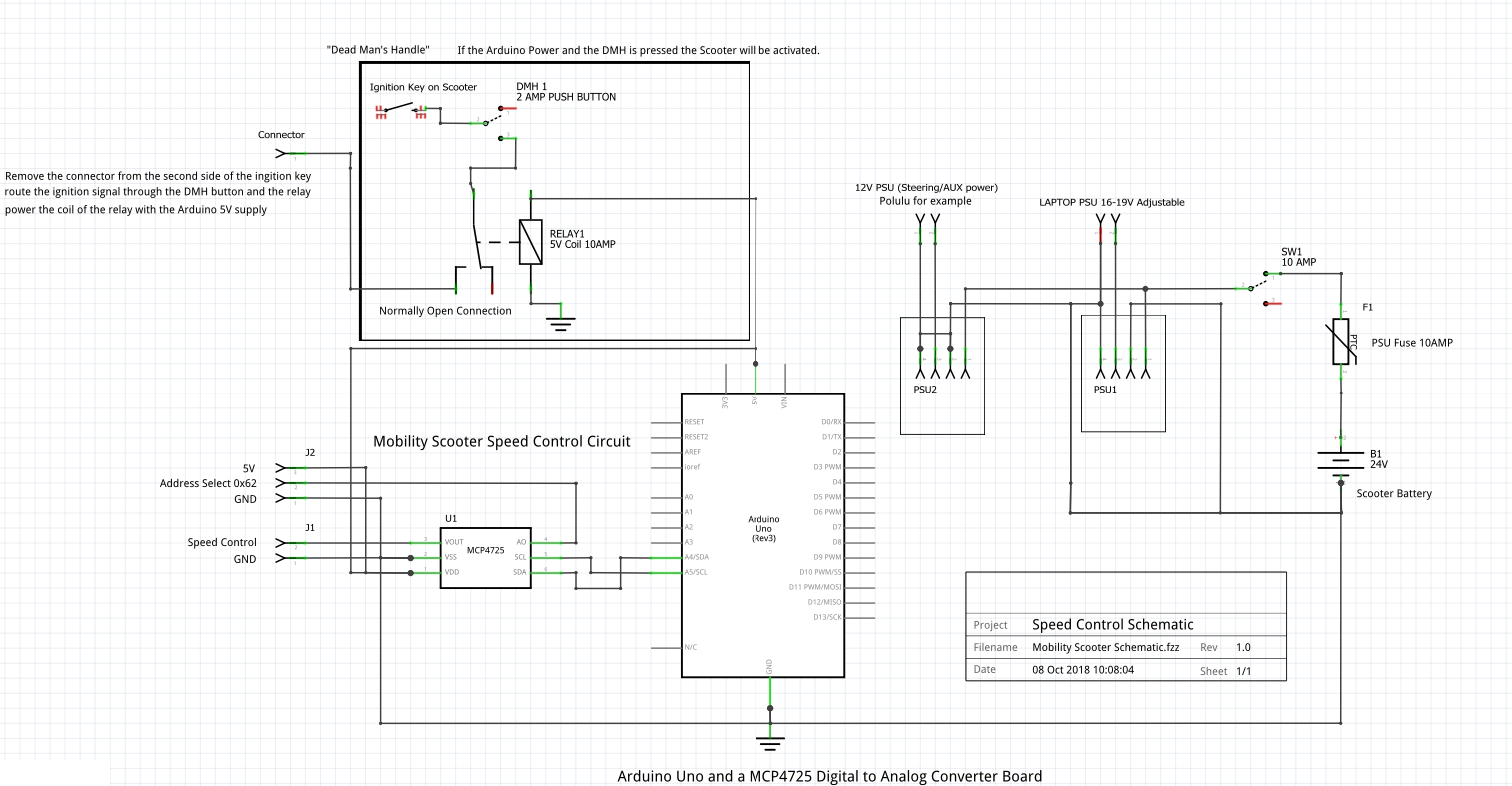}
		\caption{Circuit diagram for electronic modifications.}
		\label{fig:circuitDiagram}
	\end{figure*}

	The new vehicle electronics, which various different voltage power supplies (cf. Fig. \ref{fig:circuitDiagram}), are packaged on a single new PCB (Printed Circuit Board), as shown in Figs.~\ref{fig:pcb_design} and \ref{fig:pcb_assembled}. This is convenient as it reduces the number of small wires between the components by having them directly drawn on the board, and packages them together.
	
\begin{figure}
		\centering
		\includegraphics[width=0.6\linewidth]{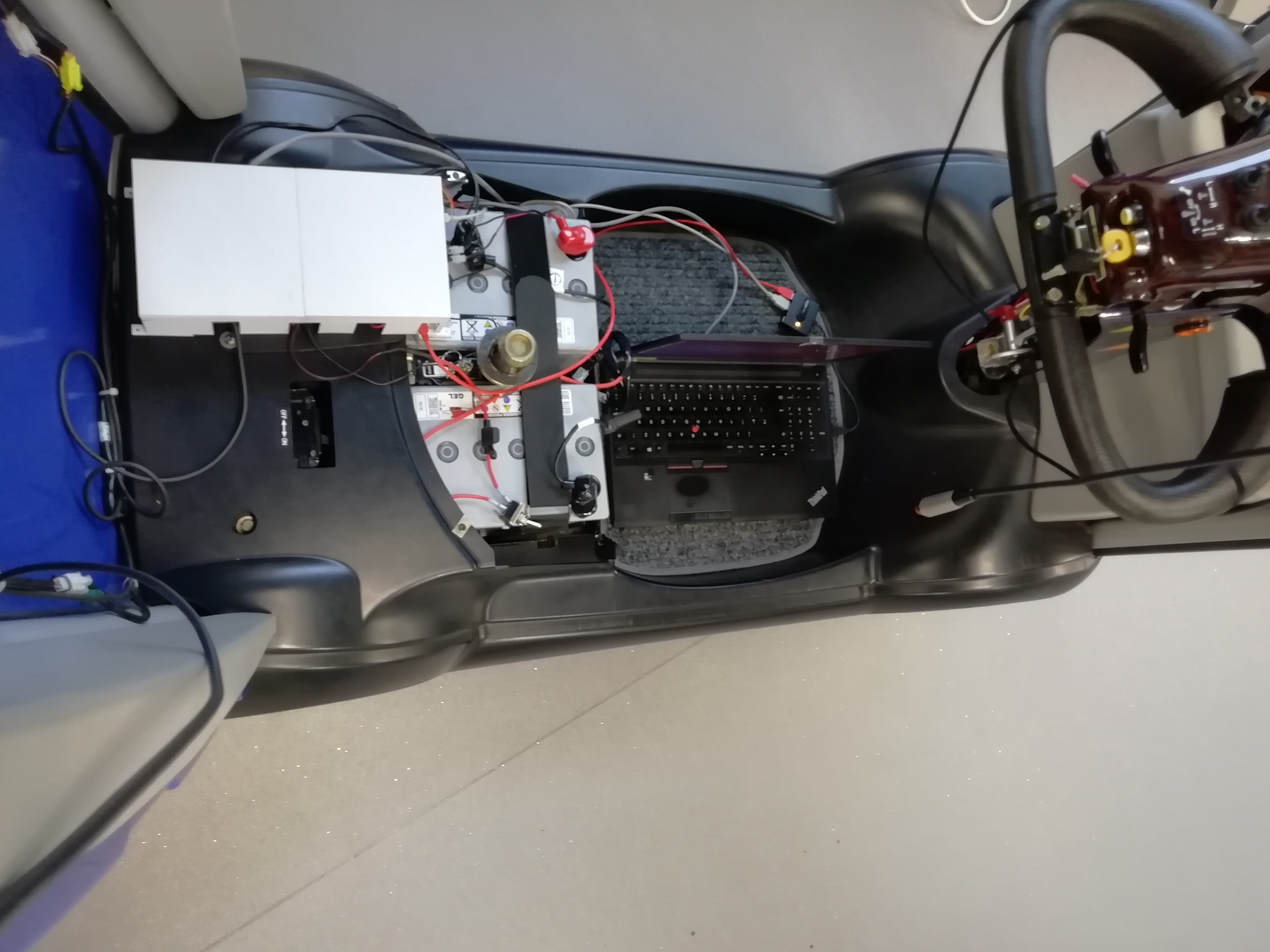}
		\caption{PCB enclosure mounted on the vehicle (top left, white box).}
		\label{fig:pcb_enclosure}
	\end{figure}

\begin{figure}
		\centering
		\begin{subfigure}{0.45\textwidth}
			\centering
			\includegraphics[width=\columnwidth]{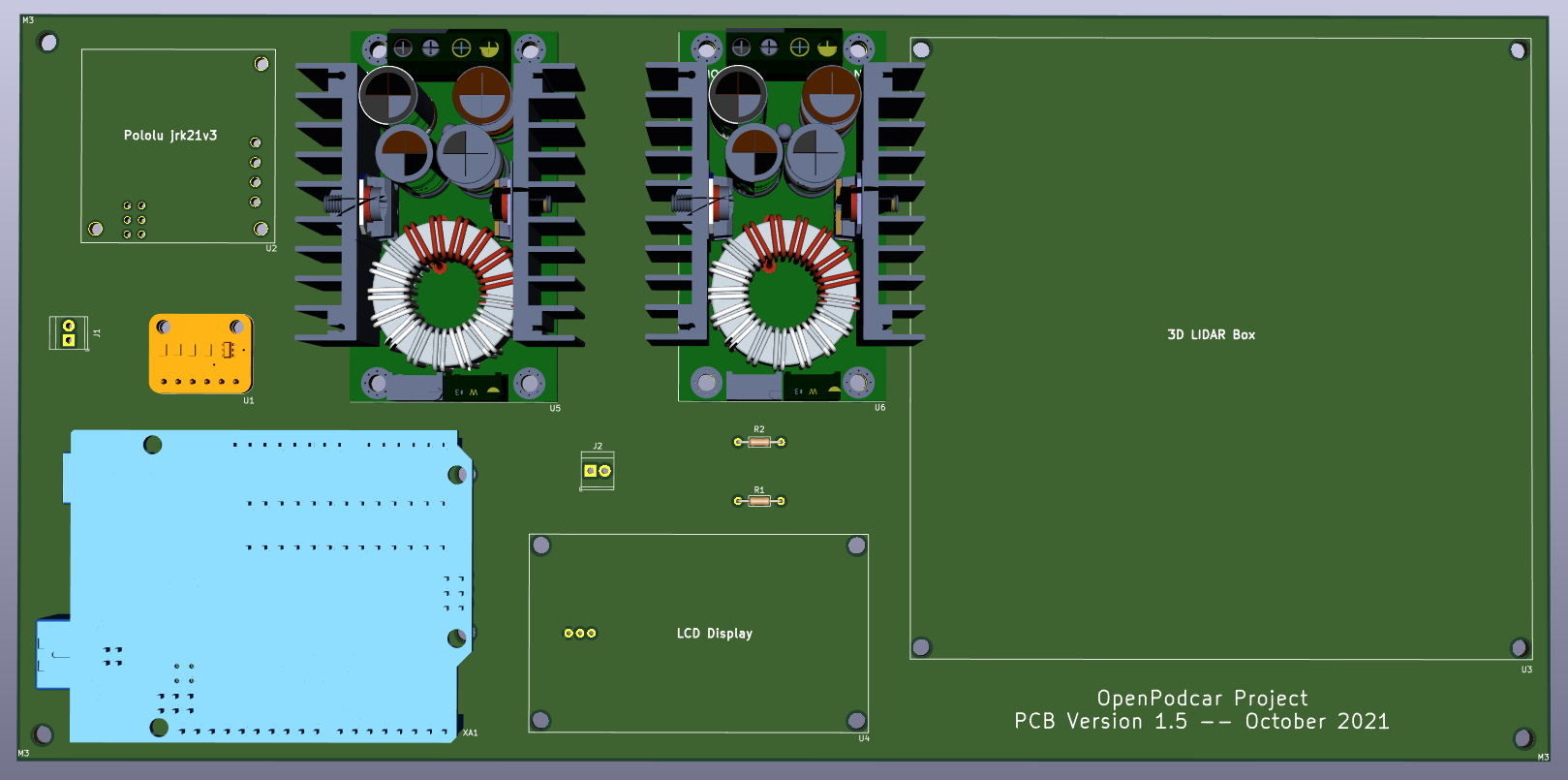}
			\caption{PCB Design.}
			\label{fig:pcb_design}		
		\end{subfigure}
		\quad
		\begin{subfigure}{0.45\textwidth}
			\centering
			\includegraphics[width=0.5\columnwidth, angle=90]{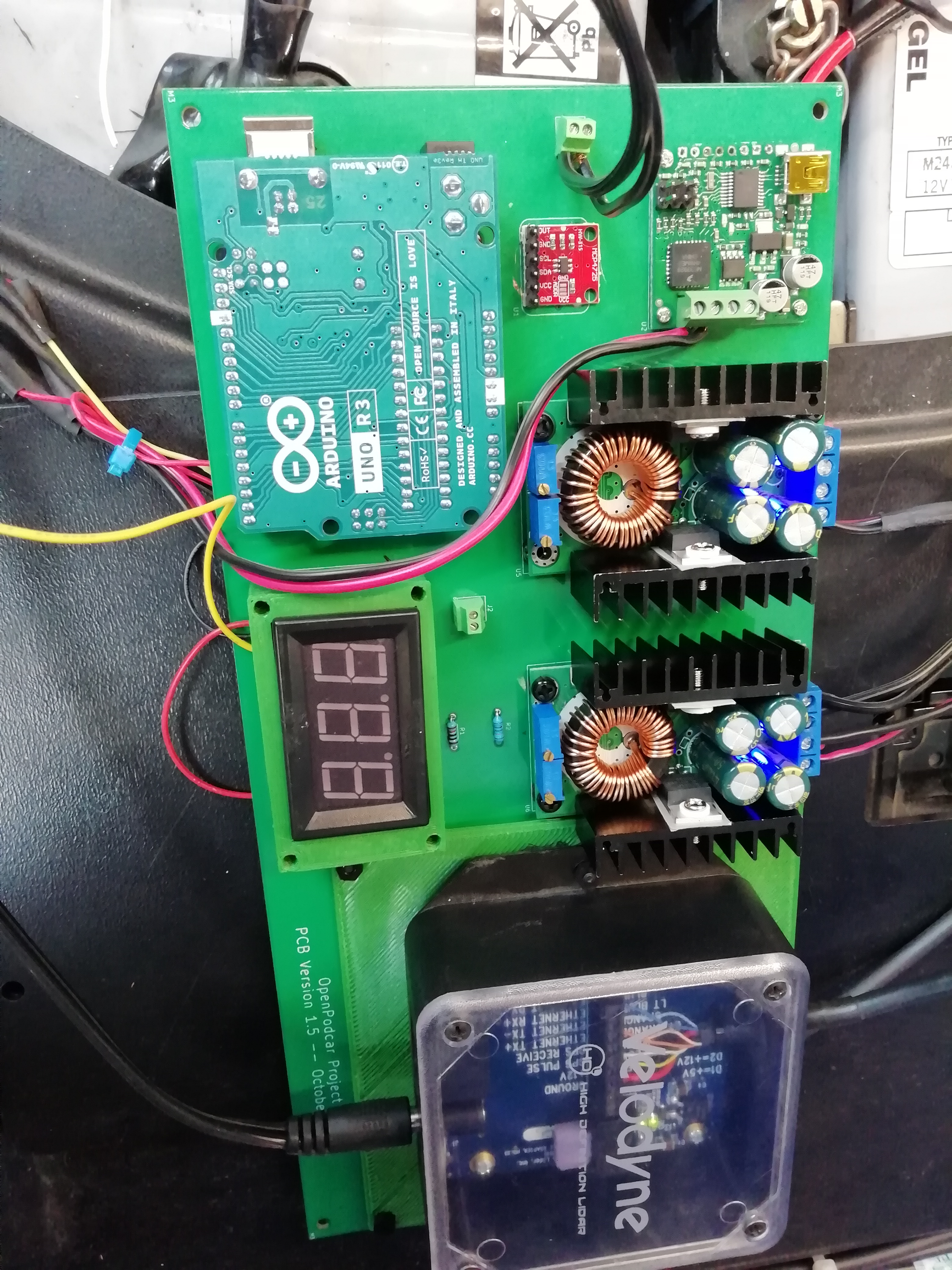}
			\caption{PCB assembly currently used.}
			\label{fig:pcb_assembled}
		\end{subfigure}	
		\caption{Electronics with PCB board design and assembly}
	\end{figure}

	As an OSH design, the PCB hosts several daughter PCBs, mounted using headers. The physical structure of the large PCB comprised of these smaller PCBs reflects the OSH design itself. There are two DC-DC Buck converters with an XL4016 regulator, an Arduino Uno, an MCP4725 DAC (Digital-Analog Converter), a Pololu Jrk 21v3 motor controller with position feedback for the linear actuator, two resistors (10k$\Omega$ and 100k$\Omega$) for the potential divider and two terminal blocks. 3D-printed parts support the mounting of the LCD and the 3D lidar to the board. A 3D printed enclosure mounts and protects the PCB board, as shown in Fig.~\ref{fig:pcb_enclosure}.
	
	\subsection{Steering Automation System}
	
	The front wheels are steered by a Pololu Jrk 21v3 PID controller-driver, which takes serial port desired positions as input. It also takes feedback position information as an analog voltage from the linear actuator as an input. It outputs analog high-power voltages to the linear actuator. A gratis, closed-source, Windows program from Pololu is required once, at build time, to set the PID parameters for the linear actuator. 
	
	The relationship between the required central turning angle $\theta$ of the pair of front wheels and extending length $l$ of linear actuator as in Fig. \ref{fig:steering} is given by,    
	\begin{gather}
	\theta = \alpha - \arctan(\frac{W}{2H}) \\
	% update W and H in figure. W is the width between two front wheels and H is the distance between front wheel and rear wheel.
	\beta = \alpha -\frac{\pi}{2}\\
	x = r_1 \cos(\beta) \\
	y = r_1 \sin(\beta) \\
	l = \sqrt{(x_0-x)^2 + (y_0-y)^2} - L + l_0    
	\end{gather}    
	where $r_1$, $x_0$, $y_0$, $W$, $H$ and $L$ are the geometric coefficients shown in Fig.1. Among them, the value of $y_0$ is negative. $l_0$ is the initial value of the linear actuator position feedback. Table \ref{tab:linear_actuator} specifies the acceptable serial port commands for the linear actuator. Sending commands outside this range may mechanically destroy the system.  
	\begin{table}
		\begin{center}
			\caption{Linear actuator acceptable command ranges.}
			\label{tab:linear_actuator}
			\begin{tabular}{ c c }
				\hline
				FA:cmd     &  Effect \\
				\hline
				2500    &    turn max right i.e $\sim$ -45 deg \\
				1900    &    center wheels i.e $\sim$ 0 deg \\ 
				1000    &    turn max left i.e $\sim$ +45 deg \\
				\hline\\
			\end{tabular}
		\end{center}   
	\end{table}

	\subsection{Speed Controller Automation System}
	
	An Arduino UNO \cite{oxer2011practical} is used to send electric signals to the vehicle's motor controller in place of the donor vehicle’s paddle controller’s potentiometer. An Adafruit MCP4725 DAC is connected to the Arduino as in Fig. \ref{fig:circuitDiagram}, and is used to send clean analog speed command voltages to the donor vehicle’s internal controller.
	
	Arduino firmware source, and upload instructions, are supplied in the repository. When uploaded to the Arduino (using the standard Arduino IDE running on the laptop), the firmware provides a simple serial port API running at 112,000 baud. It receives ASCII commands of the form `FA:210' as speed commands. Table \ref{tab:speed_commands} summarises the range of speed commands and their corresponding output voltages. 
	
	\begin{table}
		\begin{center}
			\caption{Speed commands and their corresponding output voltages.}
			\label{tab:speed_commands}
			\begin{tabular}{ l c c }
				\hline
				Command   &    Voltage            &  Effect \\
				\hline
				FA:0      &    0                  & very fast reverse  \\
				FA:80     &    $\sim 0.9$         & fast reverse (ROS limit) \\
				FA:132    &   $\sim 1.5$          & slowest reverse motion  \\
				&                       & dead zone - allows ignition \\
				FA:170    &    1.9                & stop - zero/home position \\
				FA:201    &    $\sim 2.3$         & slowest forward motion \\
				FA:240    &    $\sim 2.7$         & fast forward (ROS limit) \\
				FA:255    &    $\sim 3.0$         & very fast forward \\
				\hline\\
			\end{tabular}
		\end{center}    
	\end{table}
	
	To start the ignition, the car safety system requires the control voltage to be in the dead range. A problem is that this doesn’t correspond precisely to any fixed speed bytes, due to floating USB power level issues.  But if we pick a number solidly in the center of the dead zone, such as 164, this will work for most USB supplies. (i.e. when the vehicle’s battery is flatter, the voltages provided to USB power by it are lower. For example, we might send 164 and get 1.9V instead of the usual 2.26V.) This may result in the vehicle not starting and producing an audible warning beep instead.
	
	Also due to floating voltages from the battery, the Arduino typically receives a lower power e.g. 4.9V instead of its ideal 5V, which gets divided by the DAC value in some calculations. 
	
\begin{figure}
		\centering
		\includegraphics[width=0.5\columnwidth]{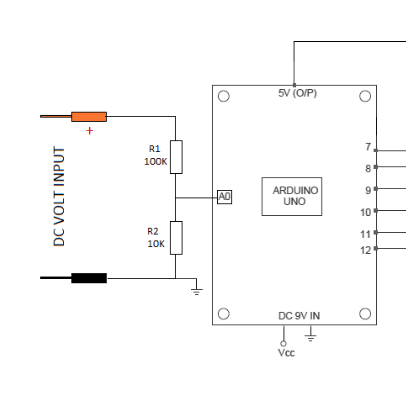}
		\caption{Potential divider linked to the battery}
		\label{fig:potential_divider}
	\end{figure}

	To deal with these instabilities, a potential divider is added to the battery to monitor its  voltage and compensate the podcar control accordingly, as in Fig. \ref{fig:potential_divider}. A ``BV'' command is provided in the Arduino serial protocol which allows callers to request this current battery voltage. This can then be used by higher-level (Python) systems to decide what speed bytes to sent, including compensating for the floating dead zone.
	
	\subsection{Software Interface (ROS)}
	
	A ROS interface to and from the physical vehicle is provided as described below. ROS is an open source operating system for robots based on a publish-subscribe pattern \cite{quigley2009ros}, which is the robotics community’s standard interface. The ROS core and software all run on a consumer laptop computer mounted on-board the vehicle, and that could be powered from a DCDC converter from the vehicle battery, running Xubuntu 16.04 (Xenial) and ROS Kinetic. 
	
	The system expects to hear two incoming ROS control messages: /speedcmd\_meterssec and /wheelAngleCmd, which contain single floats representing the desired speed in meters per second, and the desired front wheel orientation in radians respectively. These two messages are received by ROS nodes speedm2arduino and wheelAngle2Pololu, which are ROS drivers for the Arduino speed controller and the Polulo steering controller respectively. Converters from a standard ROS USB joystick driver node to the speed and angle command interface messages are provided, by joystick2speedms and joystick2wheelAngle. These use the $y$ axis of a joystick for speed and $x$ for steering.
	
	\subsection{3D Lidar Sensor}
	A Velodyne VLP-16 lidar sensor is mounted on the vehicle roof using a Manfrotto Black Digi Table Tripod 709B. It is mounted at a 10 degree tilt downwards (to allow pedestrians to be most clearly seen in the 16 scan lines). The lidar has a ROS driver. 
	
	\subsection{High-Level Automation Software}
	
\begin{figure}[h]
		\includegraphics[width=\columnwidth]{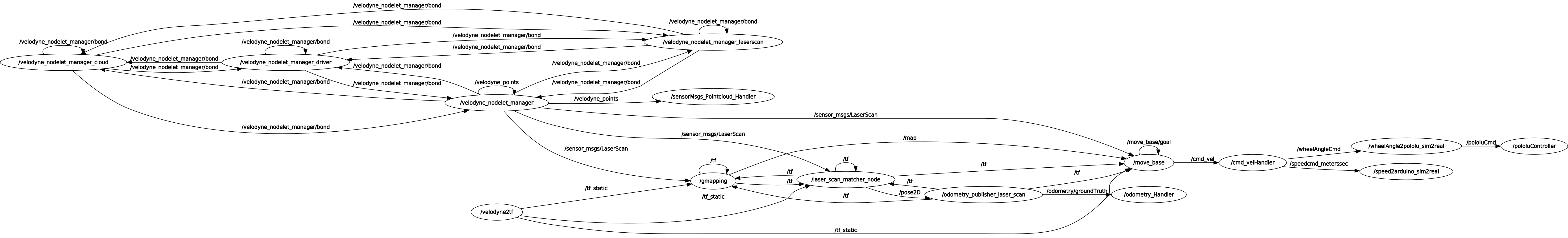}
		\caption{ROS nodes used in the autonomous driving mode.}
		\label{fig:rosnode_autonomous}
	\end{figure}

	Fig. \ref{fig:rosnode_autonomous} shows an overview of the ROS components used in high level automation, including localisation and mapping, path planning and control, and pedestrian tracking as discussed in the following sections.
	
	\subsubsection{Localisation and Mapping System}
	
	Simultaneous Localisation and Mapping (SLAM) \cite{thrun2002probabilistic} is the robotic task of inferring the robot’s location at the same time as building a map of its environment, which is a classic `chicken and egg’ problem as the two subtasks depend on one another. Solving SLAM is an NP-hard problem but many standard approximations exist. GMapping \cite{yuen2017improved} is a ROS implementation of a Rao-Blackwellized Particle Filter (RBPF) in which ``each particle carries an individual map of the environment''. The information carried by each particle overlaps, and an estimation of a map can be built based on these relationships. As the robot moves around the environment, these estimations are stored, and when a ‘feedback loop’ is closed, the estimations cascade into a portion of the completed map. These maps take the form of 2D occupancy grids, and can be used later by the navigation stack to plan paths around the environment.  To provide reliable odometry data for GMapping, ROS laser\_scan\_matcher package is used  as a stand-alone odometry estimator that matches consecutive laserscans.

	\subsubsection{Path Planning and Control}
	
	Path planning is the autonomous selection of an entire desired trajectory for a robot to get from a current pose to a desired pose. Path control (or path following) is then the real-time process of executing a path plan by interactively monitoring the robot’s state and sending commands to motors, to make the actualized path close to the desired path. The OpenPodcar software includes path planning and control with the standard ROS tool move\_base and Timed Elastic Band (TEB) \cite{rosmann2013efficient} plugin. These tools implement the requirement geometry of Dubins paths \cite{dubins1957curves} and Ackermann steering. The values for parameters such as minimum turning radius have been calculated from the technical specifications of the base vehicle \cite{shoprider2016shoprider}.
	
	\subsubsection{Pedestrian Detection and Tracking}
	
	A pedestrian detector and tracker ROS package are included in the system.  The lidar-based detections are classified by a SVM (Support Vector Machine) classifier, then a Bayesian multi-target tracker is used to track pedestrians over time. These modules re-use OSS from the EU FLOBOT project \cite{yan2020robot}, merged into the repository.

	\subsection{Simulation}
	
	\begin{figure}
		\centering
		\begin{subfigure}{0.45\textwidth}
			\centering
			\includegraphics[width=0.75\columnwidth]{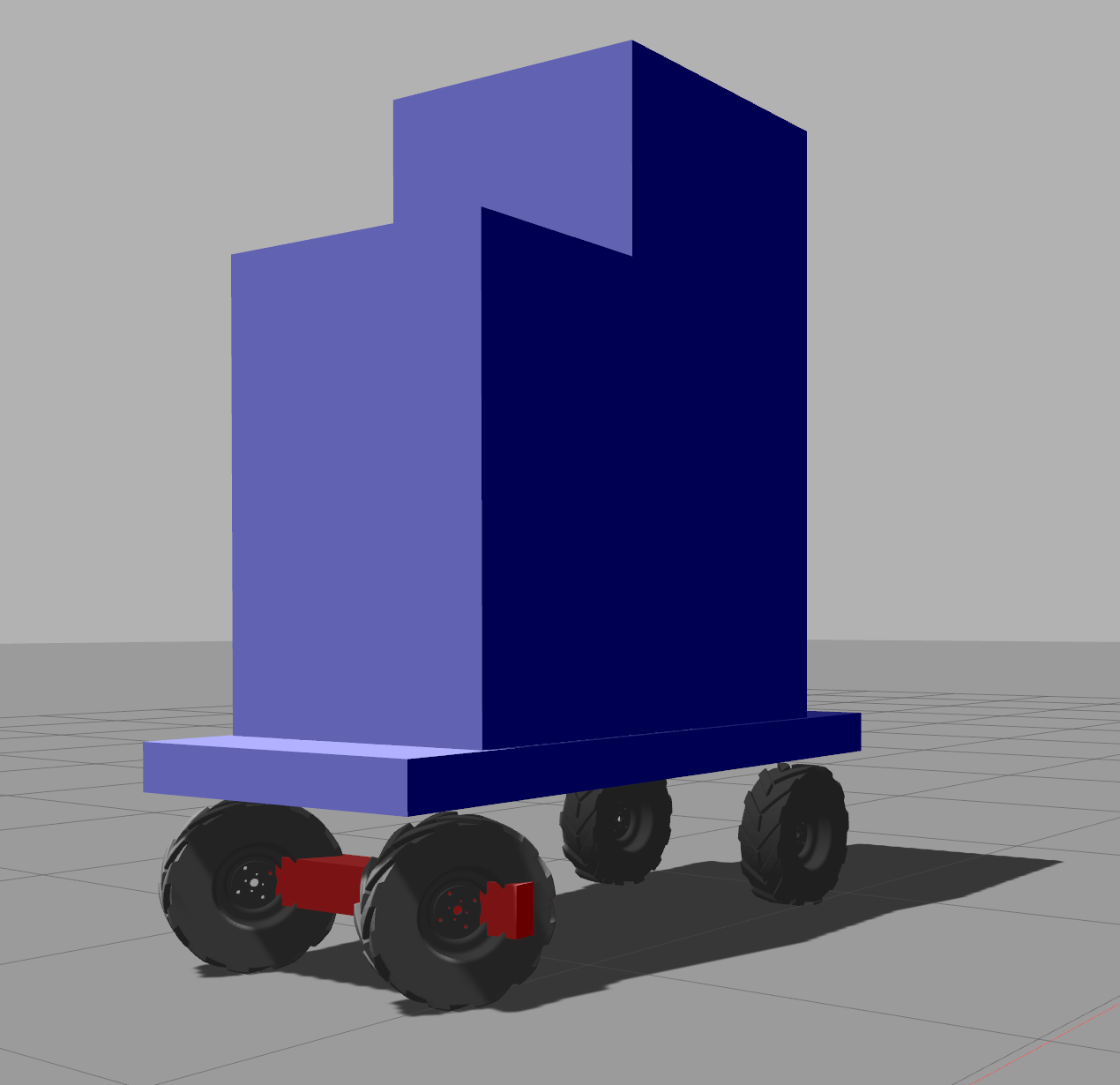}
			\caption{Physical simulation of vehicle.}
			\label{fig:physSim}
		\end{subfigure}	
		\quad
		\begin{subfigure}{0.45\textwidth}
			\centering
			\includegraphics[width=0.85\columnwidth]{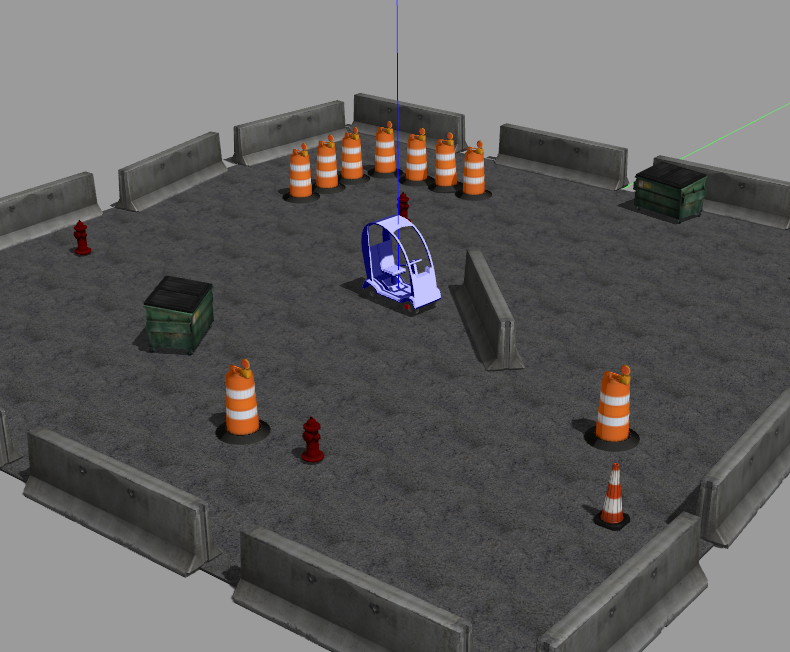}
			\caption{Default Gazebo simulation world}
			\label{fig:sim_world}
		\end{subfigure}
		\caption{OpenPodcar 3D simulation}
	\end{figure}
	
	\begin{figure}
		\centering
		\begin{subfigure}{0.45\textwidth}
			\centering
			\includegraphics[width=\columnwidth]{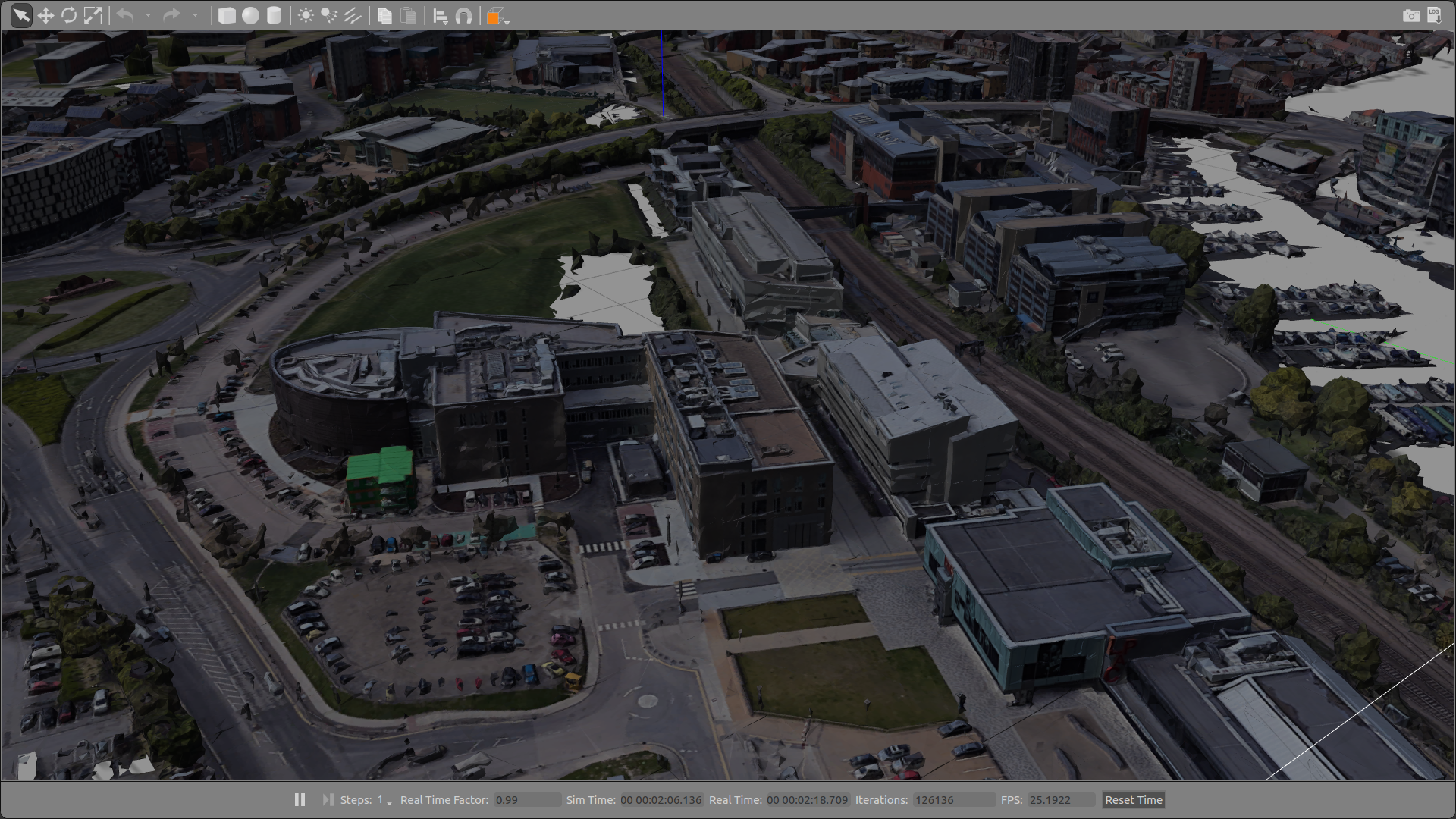}
			\caption{University of Lincoln 3D world}
			\label{fig:INB_world}
		\end{subfigure}	
		\quad
		\begin{subfigure}{0.45\textwidth}
			\centering
			\includegraphics[width=\columnwidth]{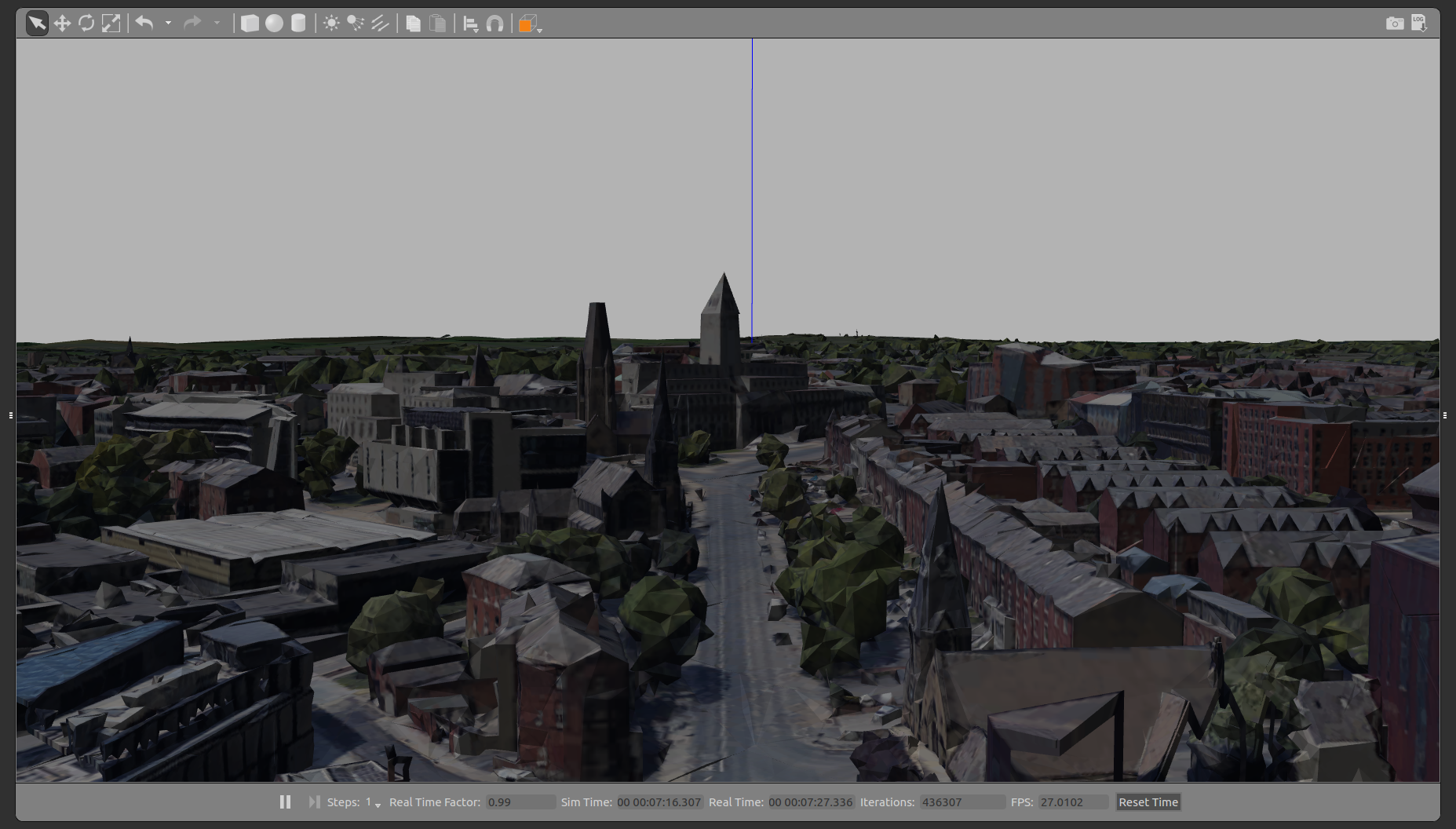}
			\caption{University of Leeds 3D world}
			\label{fig:Leeds_world}
		\end{subfigure}	
		\caption{OpenPodcar additional Gazebo 3D simulation worlds}
		\label{fig:gazebo_new_worlds}
	\end{figure}
	
	\begin{figure}
		\centering
		\includegraphics[width=0.7\columnwidth]{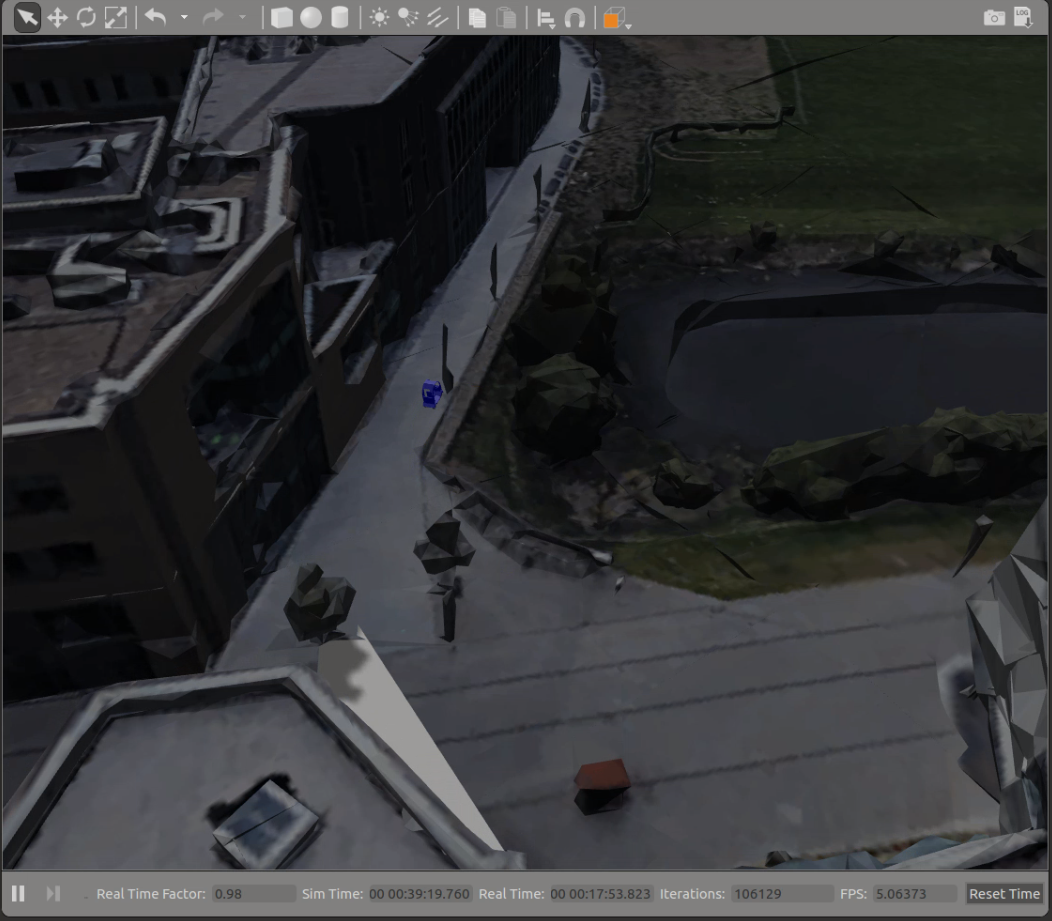}
		\caption{OpenPodcar in Lincoln world}
		\label{fig:sim_INB_world}
	\end{figure}
	
	A physical simulation of the vehicle is provided for use in Gazebo 7 \cite{koenig2004design} under ROS Kinetic and Ubuntu 16.04 (Xenial). The simulation implements the same ROS interface as the physical vehicle system to enable plug and play inter-operability between them. The physics model is based on a simplified vehicle geometry with two large cuboids containing the vehicles’ mass, as shown in Fig. \ref{fig:physSim}. Wheel geometry, friction, and motor driver parameters were measured from the physical vehicle. A detailed graphical mesh model of the vehicle is provided for display, rather than physical simulation, purposes. The main difference with the real vehicle is that the effects of the linear actuator are represented by a tracking rod, where is mounted the Kinect sensor used in place of the lidar, as found in Fig. \ref{fig:sim_world}.
	
\begin{figure}[h]
		\includegraphics[width=\columnwidth]{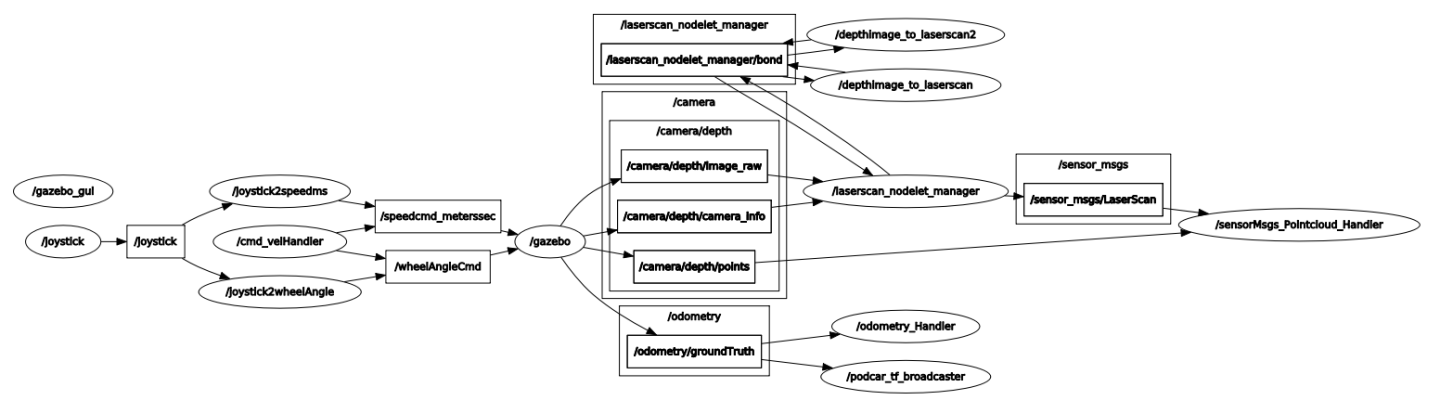}
		\caption{ROS nodes used in simulated,  manual joystick control mode.}
		\label{fig:sim_nodes}
	\end{figure}

	A basic 3D world containing the podcar and various test objects from Gazebo libraries is provided by default as shown in Fig. \ref{fig:sim_world}. Fig. \ref{fig:sim_nodes} shows the complete ROS node configuration used during simulation, under manual joystick control. Moreover, the open source Blender 3D add-on, called MapsModelsImporter \cite{michelmaps}, was used to create further 3D worlds representative of the University of Lincoln, the testing area for the OpenPodcar, and the University of Leeds campuses, shown in Fig.\ref{fig:gazebo_new_worlds}. Fig.~\ref{fig:sim_INB_world} shows the OpenPodcar in Lincoln campus environment.
	
	\section{(3) Quality Control}\label{h.f8237gmzmwc6}
	
	\subsection{Safety}\label{h.v60aduckfisj}
	
	Autonomous vehicles can present a significant hazard to humans and to the environment in which they operate. Damage to surroundings and possible injury to operators and bystanders could result from inappropriate use or malfunction. A particular risk arises from the speed controller on the donor vehicle being of 'wigwag' style, as is common in mobility scooters. This means it is an analog signal in the range 0-5V, including a dead zone around 2.5V corresponding to no motion. Above the dead zone and up to 5V are forward speed control commands of increasing speeds, below the dead zone to 0V are reverse control commands of increasing speeds. Wigwag control is  potentially dangerous because a 0V signal might appear due to component failure rather than as a desired max-speed reverse command. Also, if the vehicle batteries run low, the scaling of this signal may be altered resulting in the dead zone position floating and leading to further undesired motion. The following layered safey systems are included to fully mitigate these risks:
	
	\paragraph{Fusing} As shown in Fig.~\ref{fig:circuitDiagram}, a 10A fuse is inserted between the vehicle's original 24V battery and the switch to the new electronics. This is in addition to original fusing and other safety features provided by the donor vehicle, which all remain in tact. 
	
\begin{figure}
		\centering
		\includegraphics[width=0.5\columnwidth]{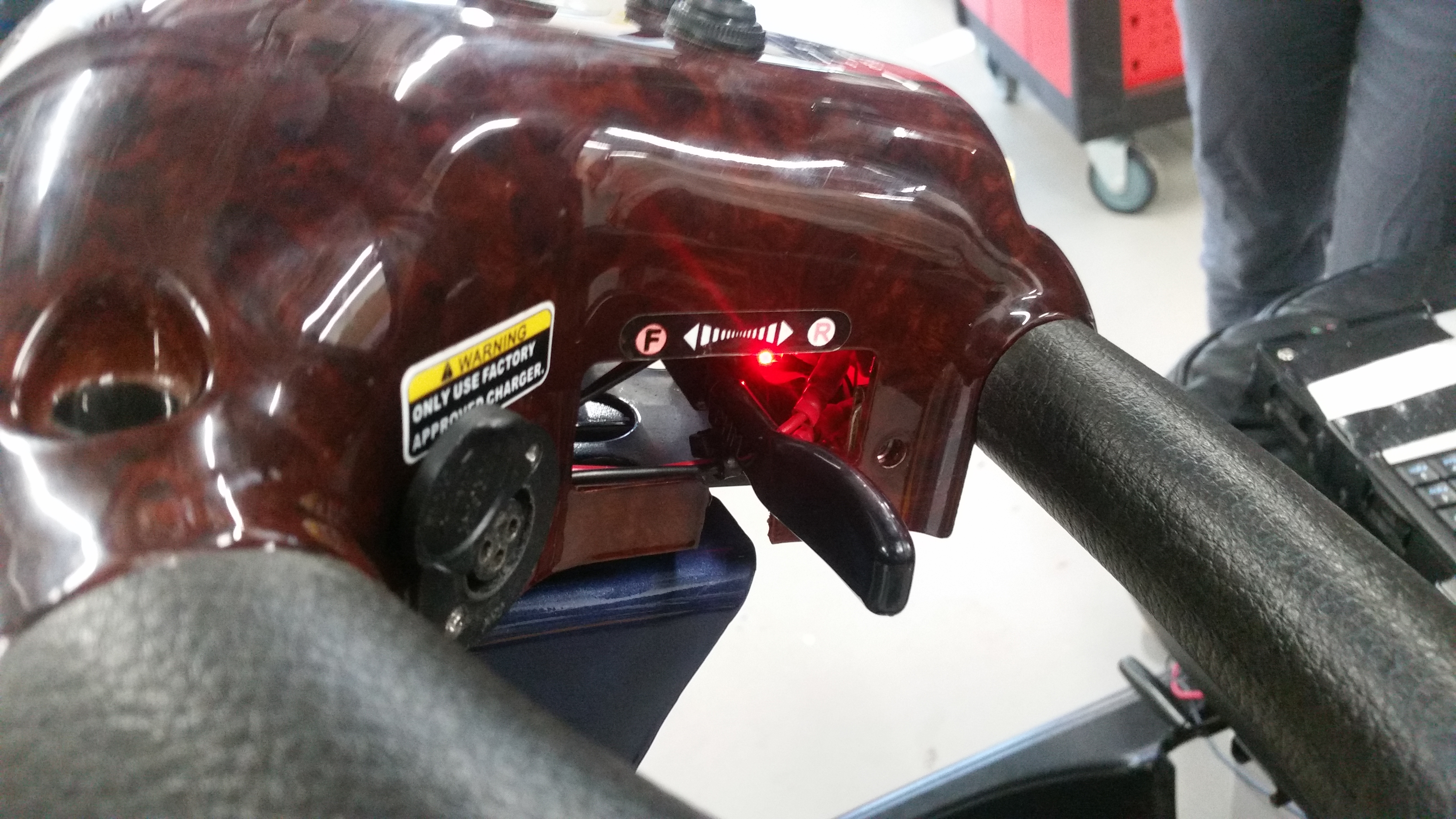}
		\caption{Steering console showing the newly added relay (with lit LED)}
		\label{fig:relay}
	\end{figure}

	\paragraph{Dead Man's Handle} It is essential that a suitable emergency stop system is implemented in all autonomous vehicles. Given the research nature of the OpenPodcar, a safety mechanism which stops the vehicle under fault conditions is an especially important part of the design. A wired dead-man's handle (DMH) is included which is required to be pressed by a human experimenter at all times, in order for a hardware relay to actively continue to supply power from the vehicle’s batteries to all other systems. The relay connects to the donor vehicle's keyed ignition switch and will naturally cut out if these signals are absent for any reason, including failures in the safety systems themselves. A photograph of the installed system is shown in Fig. \ref{fig:relay}.
	
	\paragraph{Heartbeat Signal} The serial protocol linking the Arduino to ROS includes a heartbeat signal, in which the Arduino code will shut down the motors unless a correctly formatted and timestamped serial command is received within 0.1 seconds. This requires ROS code to actively check and confirm its own status and to send positive confirmation, for example if ROS or Linux go down then this heartbeat will cut off.
	
	\paragraph{Steering Control System Limiter} Limitations are placed on the steering controller for the linear actuator commands, to only allow the vehicle to accept and execute input values within the range that will keep the mechanical mounting safe.

	\subsection{ General Testing}\label{h.wbekh9ay82yu}
	
	A series of sub-component (e.g. Pololu, DAC) acceptance tests, component (e.g. PCB, lidar) hardware unit tests, and system integration tests are defined and included as formal, non-optional steps in the build instructions.  The structure of the tests is designed to enable build problems to be immediately localised, so that passing one test means that a failure of the next one must be due to build steps that have occurred between them. Below is a summary of these tests.
	
	At component level, an external power supply, a multi-meter, a clamp meter and a breadboard with some wires are frequently used to recreate smaller electronic circuits in order to check the voltages, currents and the correct functioning of each component during the build. For instance, a circuit with an external power supplying 5V to the Arduino connected to the DAC is temporally created to test the Arduino code and its communication with the DAC.  Similarly, another test circuit is created with an external power supplying 12V to the Pololu connected to the linear actuator to send direct commands via the Windows program used to fix the PID parameters. These hardware unit tests are essential to the success of components' integration to the vehicle and make things easier later. 
	
	At system integration level, udev rules are used to facilitate testing with the creation of simlinks, i.e. dynamic assignments for the laptop USB ports connected to the Arduino and the Pololu, using their respective product and vendor IDs. This helps in being able to physically interchange the USB ports without having any impact at the software level. For the speed control, the vehicle wheels are lifted from the ground using jacks to stop them from driving off. This technique helps to test and fix the Arduino and ROS speed control code whilst staying in the same place. Vehicle steering is first tested using the Windows app that allows direct commands to be directly sent to the linear actuator. This helps verify and fix the linear actuator mounting as desired. Similarly, using Pololu's C++ API, direct commands are sent from a terminal to the linear actuator, but this time for testing at the software level. 
	
	Driving tests are initially performed in the manual joystick control mode in order to ensure that both hardware and software stack work well together. In particular, the LCD on the PCB board helps with checking in real-time the voltage received for each speed command and the LEDs colors displayed on the Pololu also give useful indications about the steering control. 
	
	The autonomous driving tests with move\_base and TEB are performed with the vehicle speed controls dial know set to `5', corresponding to about $0.2$m/s. This relatively low speed is chosen because these tests may be performed in a shared and cluttered research lab around people. Also, a large inflation distance is set in the planner to prevent the vehicle from close contacts with both static and dynamic obstacles. At first, simple and short goals are sent to move\_base such as ``drive one meter forward and keep your current orientation'' or ``drive three meters forward and keep your current orientation''. Once the vehicle is able to execute and reach these simple goals, more complex goal commands are sent. Once a goal is reached, it is possible to resend immediately another goal without having to turn off the system, which is very convenient for example when one wants to ask the vehicle to return to its starting position or go somewhere else. 
	
	Setting a very high accuracy for goals such as $1$mm and $0.01$rad is achievable on the vehicle and is tested for short drives in the lab. However, in these cases, the short drives may end up taking a lot of time, for example it can take up to three minutes to simply reach a one meter forward goal. This is due to the planner's oscillating behaviour around the goal. To fix this, more tolerance should be given for the goal accuracy, for example, 150mm and 0.15rad give an acceptable vehicle behaviour. During these driving tests, ROS topics and RViz (ROS visualization tool) are particularly monitored to get informed about the vehicle behaviour in real-time. 
	
	OpenPodcar was developed, and our own build was heavily tested, between March 2018 and March 2022. With its first automated test drives taking place since summer 2018 and an estimated 100km or more driven to date, the vehicle design has thus proven robust enough for autonomous vehicle research.

	\section{(4) Application}\label{h.f78bi3oom0mu}
	
	\subsection{Use Cases}
	
	\subsubsection{Self-Driving  Research}\label{h.4q5g9edishy3}
	
	\begin{figure}
		\centering
		\includegraphics[width=\columnwidth]{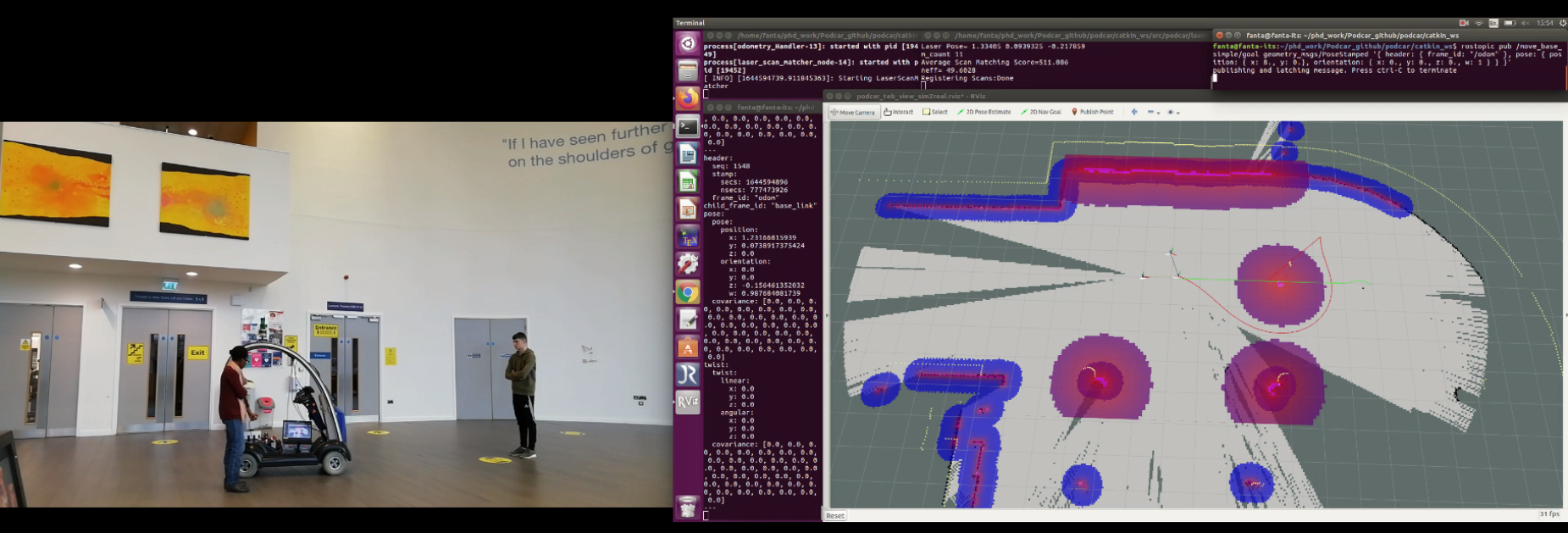}
		\caption{OpenPodcar test drive with GMapping SLAM, ROS move\_base with TEB planner and obstacle avoidance.}
		\label{fig:testDrive_gmapping}
	\end{figure}
	
	Many AV researchers cannot currently afford the acquisition of a self-driving hardware platform for their work. The OpenPodcar is primarily designed for this purpose, as a low-cost and an all-in-one, software and hardware platform for researchers and hobbyists. Thus, giving them not only the opportunity to reproduce, develop and test algorithms on a physical hardware platform but also to extend its capabilities with new features. 
	
	The \nameref{related_systems} section found that there are many open source software stacks without related open hardware platforms.  OpenPodcar thus fills this gap, offering the opportunity not only to deploy Autoware or other types of AV software but also to extend the hardware capabilities to the point where OpenPodcar could become a standard test bed for the AV research community. For example, this platform could be useful to test different SLAM and planning algorithms, parallel and valet parking methods. The objective being that both hardware and software can be tested regularly in real-world conditions and contribute towards the deployment of AVs. The OpenPodcar can avoid both static and dynamic obstacle using the integrated feature in move\_base and TEB planner. Fig. \ref{fig:testDrive_gmapping} shows the OpenPodcar test drive with GMapping, move\_base and TEB planner in action when it encounters an obstacle on its path.     
	
	\begin{figure}
		\centering
		\includegraphics[width=0.7\columnwidth]{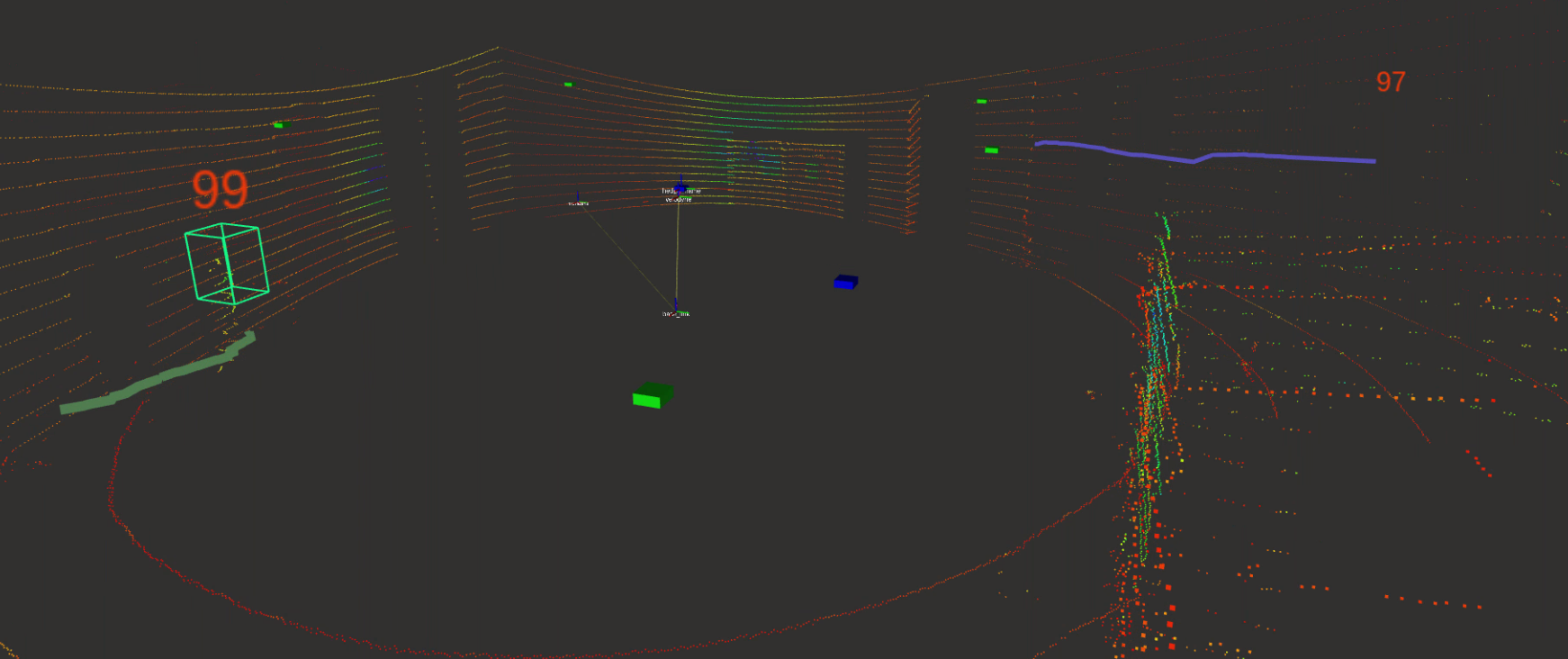}
		\caption{Pedestrian detection and tracking output from RViz.}
		\label{fig:detection_tracking}
	\end{figure}
	
	\subsubsection{Human-Robot Interaction Research}
	
	Understanding human behaviour and interaction strategies are of upmost importance nowadays for autonomous systems. There is a general growing interest from the robotics and autonomous vehicle research communities to tackle the numerous challenges posed by human interactions. Social robots as well as autonomous vehicles need better models of human behaviour \cite{camara2020pedestrian, camara2020pedestriana}. Some of the authors (FC and CF) are particularly interested in improving autonomous vehicles' decision-making using a game theoretic approach for road-crossing scenarios \cite{fox2018when}. Several empirical studies e.g. \cite{camara2018towards, camara2020continuous, camara2018empirical}, were performed in highly safe lab environments and found that human participants were not interacting realistically with the other agent. A similar experiment performed in a VR environment showed a more realistic behaviour from the participants \cite{camara2021evaluating, camara2019towards}. An additional model of human proxemics (i.e., interpersonal distances) has been developed and is being combined with the game theory model \cite{camara2020space, camaraextending}. In future work, the OpenPodcar will be used to extend these human experiments using a real physical platform and demonstrate the operation of game theoretic behaviour on a autonomous vehicle for the first time. The pedestrian detection and tracking feature will be particularly useful for this task, since the AV needs to track the pedestrian in order to make a decision. Fig. \ref{fig:detection_tracking} shows an example output of the pedestrian detection and tracking integrated in the OpenPodcar.

	\subsubsection{Practical Transportation}
	
	\begin{figure}
		\centering
		\includegraphics[width=0.5\columnwidth]{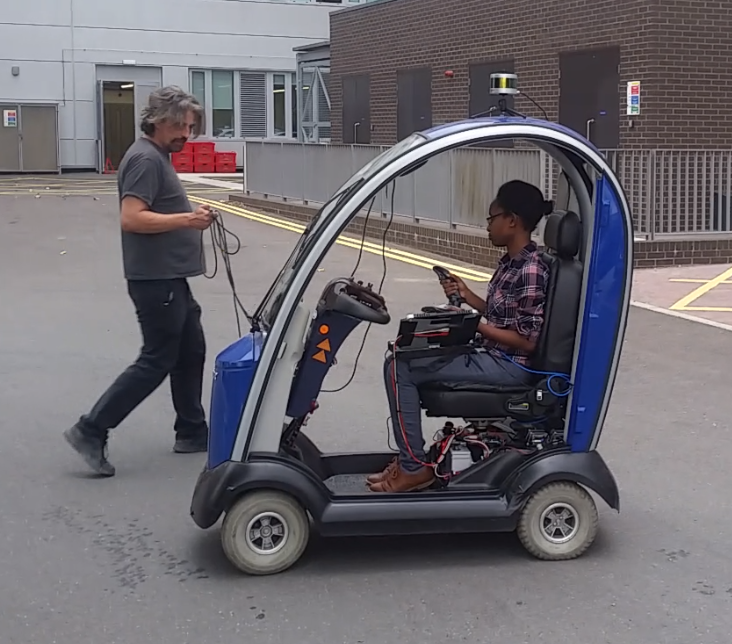}
		\caption{OpenPodcar test drive in remote control mode.}
		\label{fig:testDrive}
	\end{figure}
	
	OpenPodcar can carry at least 76kg of payload, such as a person or parcels, making it potentially useful for real-world as well as research applications.
	
	Last mile delivery of parcels could replace human workers for e-commerce deliveries. Urban center retail environments may also be improved by replacing the last mile of supply {\em to} retail outlets. Instead of driving to a shop to deliver goods, heavy goods vehicles could instead park a mile outside the urban center and transfer the goods to OpenPodcar or similar electric autonomous vehicles to take to the shop, reducing local pollution. The Covid-19 pandemic emphasised a specific need for {\em autonomous} last-mile delivery: to reduce the need for human contact and potential disease transmission at the point of delivery.
	
	OpenPodcar is able to transport a human passenger, as shown in Fig. \ref{fig:testDrive}, as it is based on an COTS mobility scooter.  For instance fleets of OpenPodcars might one day transport people over the last mile from the train station to their office, as a low cost electric taxi service. This will require more automation software to operate in busy urban environments.
	
	\subsection{Reuse Potential and Adaptability}\label{h.6wkumyl0ejrh}
	
	The OpenPodcar design is intended so that the mechanical, electronics and software components can be easily ported to other vehicles/platforms and only require small changes on the software side to adapt it and fix some parameters specific to the new vehicle requirements. This could include future deeper OSH vehicles as well as additional commercial donor vehicles. Cheaper sensors such as depth cameras or stereo cameras could be used instead of the 3D lidar. Such modifications would typically require an advanced rather than intermediate designer/builder.

	\section{(5) Build Details}\label{h.l8i9vokvs0bj}
	
	\subsection{Availability of Materials and Methods}\label{h.60suejv0jlzi}
	
	The design is made under the CERN-OSH-W licence which allows for the use of commercially available proprietary components such as the off-the-shelf donor vehicle. However the design is intended to be easily modifiable for transfer to other base vehicles, including those which are OSH at lower levels. The PCB can be manufactured by many online PCB manufacturers. The additional mechanical and electronics used are common parts available from standard online vendors. 
	
	\subsection{Ease of Build}\label{h.wg823sgyb1e4}
	
	The vehicle modification requires the use of common hand tools for assembly: spanners, screwdrivers, and pliers. Additionally, a 3D printer is needed to fabricate some components. Basic soldering skills are needed for assembling the PCB. 
	
	\subsection{Operating Software and Peripherals}\label{h.uz77dixfh5i4}
	
	The system requires open source software: Arduino IDE, Ubuntu 16.04, ROS Kinetic, Gazebo, KiCad (PCB Design), ROS GMapping, ROS move\_base, and Velodyne lidar driver. It also requires the Pololu Configuration Utility Manager software which is available gratis from the manufacturer website. The on-board laptop should have minimal specifications of amd64 3GHz quad-core, 8GB RAM, 250Gb hard-disc, USB and Ethernet ports. The system might also work on lower specifications. Step-by-step instructions for installation of these software dependencies, and the new system software components, are provided in the repository.
	
	\subsection{Hardware Documentation and Files Location:}\label{h.nbisrsde6sc3}
	
	\textbf{Archive for hardware documentation, build files and software}
	
	\textbf{Name}: GitHub
	
	%\textbf{Persistent identifier:} e.g. DOI, etc.
	
	\textbf{Project repository:} \url{https://github.com/OpenPodcar/OpenPodcar}
	
	\textbf{Licence:} CERN-OHL-W for hardware design and build instructions; GPL for software source code.
	
	\textbf{Date published:} 09/05/2022
	
	The hardware is structured as two separate formal OSH designs, each licenced as CERN-OSH-W. The first covers all components which are easily transferable to other vehicles without modification. The second contains all components which are specific to the mobility scooter donor vehicle. This structure enables the first design to be used as sub-component of closed products while also preventing closed modifications of it.

	\section{(6) Discussion}\label{h.90jl7wm65t65}
	
	\subsection{Conclusions}\label{h.h3fr33ylzsnh}
	
	OpenPodcar is a multi-purpose hardware and software platform for autonomous vehicle research. It provides the required hardware and software tools to carry out research in this field. The platform has a lower-level stack, a higher-level stack and a simulator for initial testing. It has several safety features to prevent hazards. The general testing carried on the vehicle shows a robust and safe design. Several use cases have been identified and successfully tested. OpenPodcar is open source to allow further improvements and extensions of its capabilities from the community. The replication of this work on a second and later vehicles will help identify build issues and continually improve the documentation.
	
	\subsection{Future Work}\label{h.neocsr410zj}
	
	OpenPodcar is designed to be extensible and modular, both at the hardware and software levels.  As well as improving the current design, the community is warmly invited to create forks such as replacing the mobility scooter with other donor vehicles -- including deeper OSH vehicles -- or extending the ROS stack to more complex on-road self-driving systems such as Autoware.
	
	In the current setup, the lidar has limited perception of obstacles that are too close and not as high as the lidar. This is generally fine, because people or objects would be seen before, but this can be problematic with objects such as desks and chairs that are not detected by the laserscans and can create unexpected collisions. For example, a low-cost alternative to lidar is to use a stereo camera for point cloud sensing. In this option, a StereoLabs ZedCam is mounted similarly on the vehicle roof. 
	
	The design currently uses ROS1 but the robotics community is slowly shifting to ROS2 for its security, real-time control and increased distributed processing features.  OpenPodcar could join this shift when all of its ROS dependencies have themselves completed it.
	
	The donor vehicle currently used it not itself OSH, and it would be interesting and useful to replace it with a more deep OSH vehicle.   Such vehicles would be based on OSH motor drivers and controllers such as the brushed OSMC \cite{robotpoweropen} or brushless ODrive v3.5  \cite{oriveroboticsodrive}.

	\section*{Paper author contributions}\label{h.fy8hbipy6kwe}
	
	Fanta Camara performed the physical podcar automation work including the mechanical design, printed circuit board (PCB), ROS localisation and mapping, path planning, people detector and tracker integration, Gazebo 3D simulation worlds of Lincoln and Leeds university campuses, testing, and wrote the documentation for the physical podcar and  the manuscript. Chris Waltham designed the initial electronics circuit and safety systems, wrote the Arduino code for the speed control and participated in the initial remotely-controlled driving test. Grey Churchill developed the podcar simulator with path planning in ROS/Gazebo and wrote the documentation for the simulator. Charles Fox supervised the work, wrote the manuscript, and wrote some ROS code. The manuscript was improved by comments from all the co-authors.

	\section*{Acknowledgements}\label{h.gu3yyarx72d6}
	
	The authors would like to thank Jacob Lord for creating the vehicle graphical mesh model, Yao Chen for scoping Dubins path methods and the mechanical design, Gabriel Walton for scoping simulation tools, Yicheng Zhang for helping with the 3D printer and many other useful tools for the PCB board, Zak Burrows for the PCB enclosure design and assistance during some autonomous driving tests.

	\section*{Funding statement}\label{h.4u1a7tugh2om}
	
	This project has received funding from EU H2020 grant 723395 interACT, and from InnovateUK  grant 5949683 C19-ADVs.

	\section*{Competing interests}\label{h.q1j1rznb43fl}
	
	The authors declare that they have no competing interests.
	
	\bibliographystyle{plain}
	\bibliography{OpenPodcar}

	\section*{Copyright notice}\label{h.jm5gcqv4g8x0}
	
	\copyright~2022 The Authors. %This is an open-access article distributed under the terms of the Creative Commons Attribution 4.0 International License (CC-BY 4.0), which permits unrestricted use, distribution, and reproduction in any medium, provided the original author and source are credited. See http://creativecommons.org/ licenses/by/4.0/.

\end{document}